\documentclass{article}
\usepackage{ijcai22}

\usepackage{times}

\usepackage{soul}
\usepackage{url}
\usepackage[hidelinks]{hyperref}
\usepackage[utf8]{inputenc}
\usepackage[small]{caption}
\usepackage{subcaption}
\usepackage{graphicx}
\usepackage{float}
\usepackage{multirow}
\usepackage{amsmath}
\usepackage{amsfonts}
\usepackage{booktabs}
\usepackage{boldline}
\usepackage[table]{xcolor}
\usepackage{pifont}
\usepackage{lipsum}
\usepackage[noabbrev, capitalise]{cleveref}

\newfloat{figtab}{htb}{fgtb}
\makeatletter
  \newcommand\figcaption{\def\@captype{figure}\caption}
  \newcommand\tabcaption{\def\@captype{table}\caption}
\makeatother

\newcommand{\leone}[1]{\textcolor{black}{#1}}
\newcommand{\haohan}[1]{\textcolor{black}{#1}}
\newcommand{\whh}[1]{\textcolor{black}{#1}}

\title{Iterative Few-shot Semantic Segmentation from Image Label Text}

\author{
    Haohan Wang$^{1*}$ \and Liang Liu$^{2}$\thanks{Equal contribution.} \and Wuhao Zhang$^2$ \and Jiangning Zhang$^2$ \and Zhenye Gan$^2$ \and \\Yabiao Wang$^{2\dag}$ \and Chengjie Wang$^{2\dag}$ \And Haoqian Wang$^1$\thanks{Corresponding authors.} \\
    \affiliations
    $^1$Shenzhen International Graduate School, Tsinghua University \\
    $^2$Tencent Youtu Lab \\
    \emails
    wang-hh20@mails.tsinghua.edu.cn,
    \{leoneliu, wuhaozhang, vtzhang, wingzygan, caseywang, jasoncjwang\}@tencent.com,
    wanghaoqian@tsinghua.edu.cn
}

\begin{document}
\newcolumntype{C}[1]{>{\centering\arraybackslash}p{#1}} %

\maketitle

\begin{abstract}
Few-shot semantic segmentation aims to learn to segment unseen class objects with the guidance of only a few support images. Most previous methods rely on the pixel-level label of support images. In this paper, we focus on a more challenging setting, in which only the image-level labels are available. We propose a general framework to firstly generate coarse masks \whh{from image label text with the help of the powerful vision-language model CLIP, and then refine the mask predictions of support and query images iteratively and mutually. During the refinement, we design an Iterative Mutual Refinement (IMR) module to adapt to the varying quality of the coarse support mask.} Extensive experiments on $\text{PASCAL}\textit{-}5^i$ and $\text{COCO}\textit{-}20^i$ datasets demonstrate that our method not only outperforms the state-of-the-art \haohan{weakly supervised} approaches by a significant margin, but also achieves comparable or better results to recent supervised methods. Moreover, our method owns an excellent generalization ability for the images in the wild and uncommon classes. Code will be available at \url{https://github.com/Whileherham/IMR-HSNet}.
\end{abstract}

\section{Introduction}
\label{intro}

In recent years, semantic segmentation has reached remarkable progress. However, a large amount of data is required for training such a task to segment unseen class objects. It differs from the human perception that a human is able to handle novel objects after learning from only a few samples.

\begin{figure}[t]
	\centering
	\includegraphics[width=0.98\columnwidth]{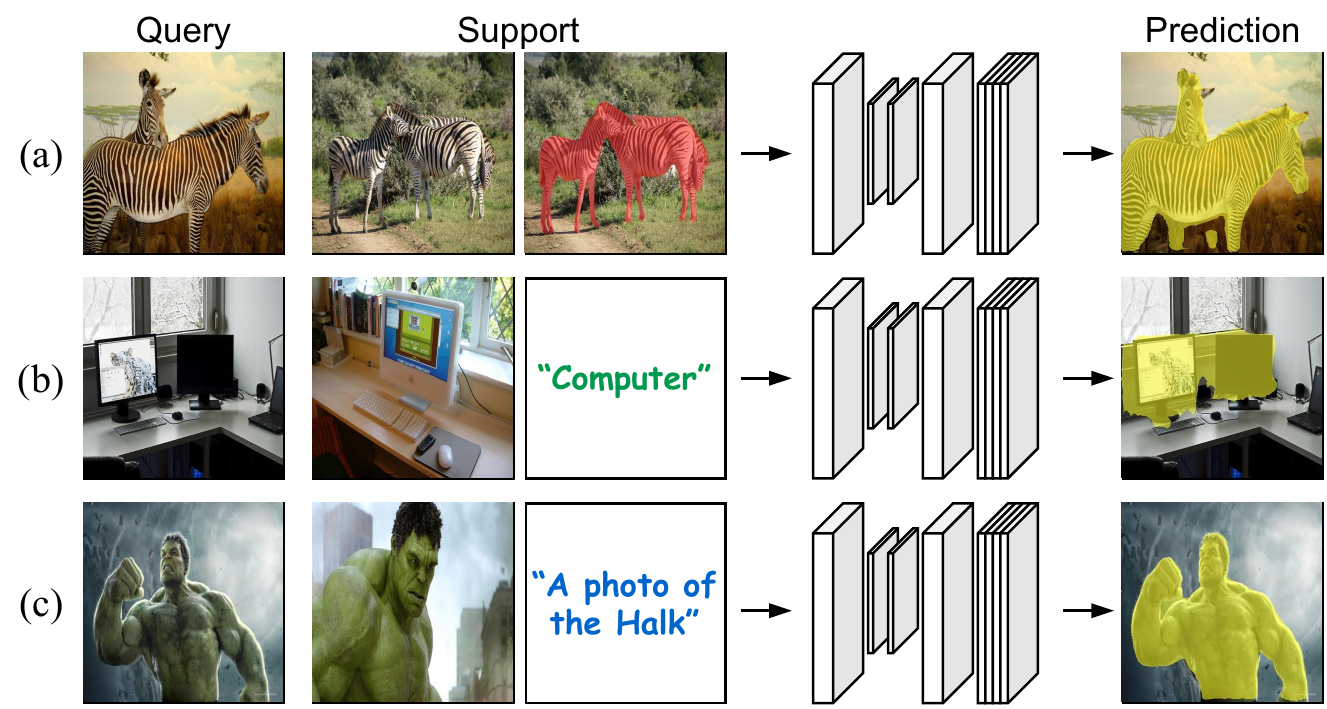}
	\caption{Illustration of (a) regular few-shot segmentation, (b) previous weakly supervised method, and (c) our method. We replace the simple class word used in previous work with a sentence of label text. Joint modeling vision and language contents and specially designed \haohan{refining} module enable our method to handle fine-grained novel classes with higher precision.}
	\label{fig:f1}
\end{figure}

Few-shot semantic segmentation is proposed to mimic the property of humans. The segmentation model is expected to segment objects with an unseen class in the query image with a few support images and pixel-wise annotations, as shown in~\cref{fig:f1} (a). Although it is more like humans and helps reduce the work of data annotation, pixel-wise mask annotation is still unavoidable. In general, compared with image-level or box-level annotation, generating the precise pixel-level label is more time-consuming. \haohan{Therefore, once faced with unseen categories, it takes much time to prepare the dense annotations, which restricts the practical application in the real-time open world, such as robotic engineering or consumer electronics products. }\par

In this work, we concentrate on the task of few-shot segmentation with image-level label support, namely weakly supervised few-shot semantic segmentation, as depicted in~\cref{fig:f1} (b). We argue that pixel-wise examples are not necessary for pixel-wise recognition, as humans are able to imagine the shape and structure of objects with the help of prior knowledge such as attribute and low-level feature. Therefore, it is meaningful to study the weakly supervised setting.

However, due to the severer data scarcity, \textit{i.e.}, the lack of precise support mask, weakly supervised few-shot semantic segmentation is rarely explored. \cite{raza-wfss} simplify the problem into the condition where only one novel class exists in the support image, but the model would collapse in the more complex scene. Recent works employ the Word2Vec embedding to associate the image feature with the class label explicitly \cite{Lee_2022_WACV} or implicitly \cite{ijcai-wfss}. Nevertheless, the performance still falls far behind that in the standard few-shot setting.\par

We summarize that the performance gap is mainly because a model designed for weakly supervised few-shot segmentation \haohan{is faced with} two daunting challenges. \textbf{\textit{The first challenge is to locate specific objects in the support image}}. Given a novel class label, the model has to filter out the corresponding objects of unseen categories from the background and other foreground instances in the support images. It always results in an imprecise or even wrong support mask, rather than the ground-truth mask. Therefore, \textbf{\textit{another challenge is to learn from imprecise labels}}. In other words, the model has to adapt to the imprecise input \haohan{support} masks.\par

In this work, we propose a two-stage framework to address the challenges. In the first stage, we tackle the first challenge, \textit{i.e.}, locating specific objects guided by the image label text. We fall back on the recent progress of vision-language model CLIP \cite{clip}, which is trained with massive image-text pairs from the web in a self-supervised way. CLIP encodes images and texts into the joint embedding space, and thus it builds a strong correlation between them. To leverage its potential, we inject the class name into a simple prompt template, as shown in~\cref{fig:f1} (c), and compute the cosine similarity of the output from vision and text encoder in CLIP. This process could be regarded as the classification task, and hence we can generate the support mask according to the high-response regions of its Grad CAM \cite{gradcam}. \par

Then in the second stage, we aim at handling the second challenge, \textit{i.e.}, the imprecise input support mask. We experimentally find that due to the robustness of the well-trained few-shot segmentation model, when segmenting the query mask with an imprecise support mask, the quality of the predicted query mask is much higher than that of the input support mask. Based on this observation, we propose to refine the support and query mask iteratively and mutually. Specifically, we continue to refine the support mask with the help of a higher-quality predicted query mask, and then update the query mask with the better support mask in turn. However, due to the varying quality of the input object masks, the regular few-shot segmentation model would struggle in judging the confidence of the input mask, which leads to a suboptimal result. Hence we propose \whh{a progressive mask refinement module Iterative Mutual Refinement (IMR)}, to capture and adapt to the varying input distribution, \textit{i.e.}, how well the input mask has been estimated. \whh{During the refinement, IMR provides an effective strategy to leverage the information of all past steps with memory, so that the support and query mask could be optimized adaptively.} We incorporate the IMR into the pipeline of few-shot segmentation, so as to better explore the potential of the iterative refinement.
Our contributions are summarized as follows:

\begin{itemize}
\item

\haohan{We propose a two-stage framework for weakly supervised few-shot segmentation, consisting of initial mask generation from the image label text and iterative support and query mask refinement.}

\item
We present an \haohan{Iterative Mutual Refinement module} to progressively refine the coarse masks, which is better for the case where the input mask quality varies. It is designed to be model-agnostic, and could serve to extend common few-shot segmentation models to the weakly supervised setting with minor modifications.

\item
Extensive experiments show that our method could not only outperform the state-of-the-art weakly supervised counterparts by a significant margin, but also close to or beyond recent regular few-shot segmentation methods. %
\end{itemize}

\section{Related Work}
\paragraph{Few-shot Semantic Segmentation.}
The dominant approaches of few-shot segmentation are devised based on meta-learning. The information of support images is aggregated into the query feature for further prediction in different ways. Most methods achieve this by first extracting single \cite{pfenet,gnn-fss} or multiple \cite{self-guided-fss,adaptive-fss} class \haohan{prototypes} from support feature by masked average pooling, then concatenating the \haohan{prototypes} and query feature as the input of a well-designed decoder. However, this simple strategy is faced with severe support information loss unavoidably. Some recent works alleviate this by exploring the potential of the pair-wise feature correlations between support and query feature, and the more powerful transformer architecture \cite{transformers-eth-fss} or 4D convolutions \cite{hsnet} is thus utilized to boost the performance further. Despite these signs of progress, few-shot segmentation requires the precise support mask during testing, which is superfluous and hardly available in customer-oriented real-time applications. In this work, we tackle this problem by replacing the ground-truth mask with the image label text.

\paragraph{Weakly Supervised Few-shot Segmentation.} Weakly supervised few-shot segmentation is rarely explored due to the severer challenge of data scarcity. \cite{panet,canet} utilize bounding box as the annotation of support images to segment the query images during testing. However, their performance is suboptimal as these models are not specifically designed for the weakly supervised condition. The first few-shot segmentation method with image-level support annotation is proposed by \cite{raza-wfss}, where they generate the pseudo masks of support images by retaining the pixels not classified as background. Nevertheless, it cannot deal with the support images containing more than one novel class. Recently, \cite{ijcai-wfss} incorporate the semantic word embedding of the novel category into the visual feature, and exploited a multi-stage attention mechanism to predict the query mask. \cite{Lee_2022_WACV} regard the pseudo Classification Activation Map (CAM) of the unseen classes as the support mask \whh{and ground truth query mask during training}, which is estimated by the CAMs of semantically similar categories. However, the performance of these approaches is far away from that using a precise support mask, which restricts its practical application. On the contrary, our method could perform close to or even beyond the state-of-the-art few-shot segmentation models, and show strong generalization ability for images in the wild.\par

\begin{figure*}[t]
	\centering
	\includegraphics[width=0.95\textwidth]{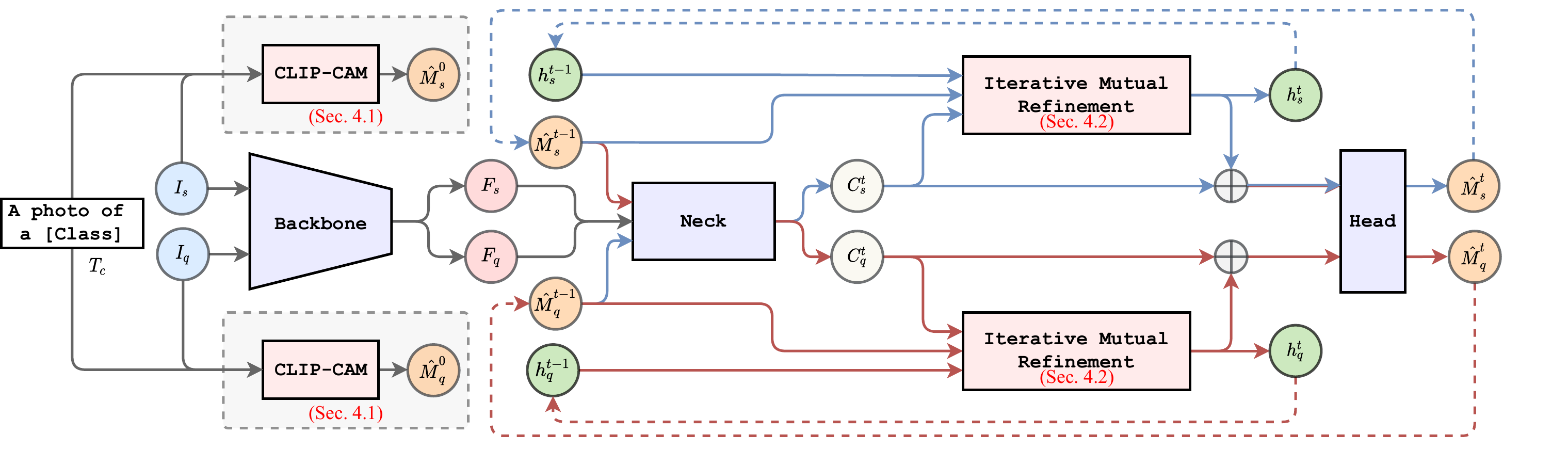}
	\caption{Pipeline of our weakly supervised few-shot segmentation method. Given a support image $I_s$ and a query image $I_q$ for a specific class $c$, we estimate coarse masks for support and query with the label prompt $T_c$ as the initial masks firstly. Then, an iterative refinement with a symmetrical data flow is used to improve the segmentation quality gradually. Note that ``Backbone", ``Neck", and ``Head" are abstract from the regular few-shot segmentation methods so that our method is able to integrate into most of the previous well-designed networks.}
	\label{fig:pipeline}
\end{figure*}

\paragraph{CLIP in Classification and Segmentation.}
 CLIP~\cite{clip} is a milestone of vision-language pretraining model, which learns a multi-modal embedding space by jointly encoding images and texts from the web in contrastive learning. CLIP shows strong capacity in zero-shot or few-shot recognition \cite{clip-adapter,clip}, and recent works have focused on the extension into dense prediction tasks like semantic segmentation. \cite{denseclip} first employ CLIP in fully supervised semantic segmentation in a model-agnostic way, and \cite{denseclip-zero-shot} extend CLIP into zero-shot segmentation. \cite{anonymous2022languagedriven} combine CLIP with a well-designed transformer to further improve the precision of segmentation. However, these methods rely heavily on complicated prompt engineering, and the upper bound of their segmentation performance is limited due to the lack of support images. Therefore, in this work, we explore the potential of CLIP in few-shot segmentation for the first time with a simple text prompting only, along with an \haohan{Iterative Mutual Refinement module} to obtain \haohan{the} precise prediction.\par

\section{Problem Definition}

Few-shot semantic segmentation aims at training a model which can segment unseen class objects given extremely limited annotated samples with the same class. Formally, the dominant episode based meta-learning paradigm employs a \textit{base} dataset $\mathcal{D}_\text{base}$ with category set $\mathcal{C}_\text{base}$ for training and a \textit{novel} dataset $\mathcal{D}_\text{novel}$ with unseen category set $\mathcal{C}_\text{novel}$ for testing. During training, a large number of  episodes are sampled from $\mathcal{D}_\text{base}$. Each episode consists of a \textit{support} set $\mathcal{S}$ and a \textit{query} set $\mathcal{Q}$ for a specific class $c \in \mathcal{C}_\text{base}$ as follows: 
\begin{equation}
\mathcal{S}=\left\{\left(I_s^{k}, M_s^{k}\right)\right\}_{k=1}^{K}, \quad \haohan{\mathcal{Q}=\left\{\left(I_q, M_q\right)\right\}}.
\end{equation}
where $I_*$ and $M_*$ denote image and mask respectively. After training, the model $f(.,\theta)$ learned to predict a binary mask $\hat{M}_q $ for query image $I_q$ with the guidance of $K$ support images and masks for a specific novel class $c\in\mathcal{C}_\text{novel}$,
\begin{equation}
\hat{M}_q = f\left(\left\{\left(I_s^{k}, M_s^{k}\right)\right\}_{k=1}^{K}, I_q, \theta \mid c\right).
\end{equation}

In the standard few-shot segmentation setting~\cite{hsnet,pfenet}, both support masks $M_s$ and query masks $M_q$ for class $c$ are available for training, while query masks are inaccessible for testing. On the contrary, we focus on a more challenging setting, named weakly supervised few-shot segmentation, in which even support masks are inaccessible in testing. Thus, the prediction becomes:
\begin{equation}
\hat{M}_q = f\left(\left\{I_s^{k}\right\}_{k=1}^{K}, I_q, \theta \mid c \right).
\end{equation}

Although a weaker annotation replaces the pixel-wise annotated mask, we show that it can still reach a comparable or even better performance.

\section{Method}

We propose a framework for weakly supervised few-shot segmentation. 
The pipeline is shown in~\cref{fig:pipeline}, which contains a CLIP-CAM module to generate coarse masks for support and query images~(\cref{sec:mask_generation}), an Iterative Mutual Refinement module to refine the coarse masks progressively~(\cref{sec:refinement}), and some basic blocks from previous well-designed models. Notably, our framework is plug-and-play and model-agnostic, and we show how to extend regular few-shot segmentation models to the weakly supervised setting in~\cref{sec:architecture}.

\subsection{Coarse Mask from Label Text}
\label{sec:mask_generation}

In the weakly supervised setting, dense annotations of supported images are not available, so we propose to generate segmentation maps as an alternative from images and class labels only. To some extent, this subtask degrades into zero-shot segmentation, for which the goal comes to segment novel objects in an image with the class label only. \whh{Nevertheless, most existing zero-shot approaches require complex architectures and training strategies, which might be inefficient to serve as only one stage of our whole pipeline.}

Recent vision-language models, especially CLIP\cite{clip}, have shown surprisingly powerful performance in associating visual and text concepts. CLIP consists of a vision encoder $E_v(\cdot)$ and a text encoder $E_t(\cdot)$, where the image and text are encoded into $f_v, f_t\in \mathbb{R}^{1 \times 1 \times \textit{C}}$. Since it learns a joint embedding space for images and texts, one can simply compute the cosine similarity of $f_v$ and $f_t$ for image-level zero-shot classification \whh{without extra training and modules.} In this work, we extend it to pixel-level zero-shot segmentation \whh{in a training-free way.} 

The original architecture of CLIP is not suitable for dense prediction tasks, so a naive idea is to adjust the architecture of the vision encoder $E_v(\cdot)$. Specifically, \whh{similar to \cite{denseclip-zero-shot}}, we drop the \textit{query} and \textit{key} embedding layers, and reformulate the \textit{value} embedding layer and the last linear layer into two convolutional layers, so as to project the image feature of each position to the embedding space of the text feature. After that, we can conduct pixel-level prediction by computing the cosine similarity between text feature and dense visual feature.

An alternative is the heatmap of Grad CAM \cite{gradcam}, which aims at looking for the region contributing most to the classification. Since CLIP computes the cosine distance between $f_v$ and $f_t$, \haohan{it} could be regarded as the classification output of a certain category, where $f_t$ works as the weight of the classifier. Thus we utilize the heatmap output by Grad CAM as the coarse object mask, as the region with a high response makes $f_v$ resemble $f_t$. 

With the guidance of the class label text, both strategies above could generate the coarse object masks from the images. We select the latter one due to its better performance. Please refer to the supplementary materials for more details of the implementation and comparison. It is worth mentioning that it is unnecessary to pursue a precise initial mask, since we adopt an iterative strategy to refine the estimation of the mask. Therefore, we do not depend on the complicated prompt engineering and other various zero-shot methods, and still achieve \haohan{a} satisfying performance.

\begin{figure}[t]
	\centering
	\includegraphics[width=.95\columnwidth]{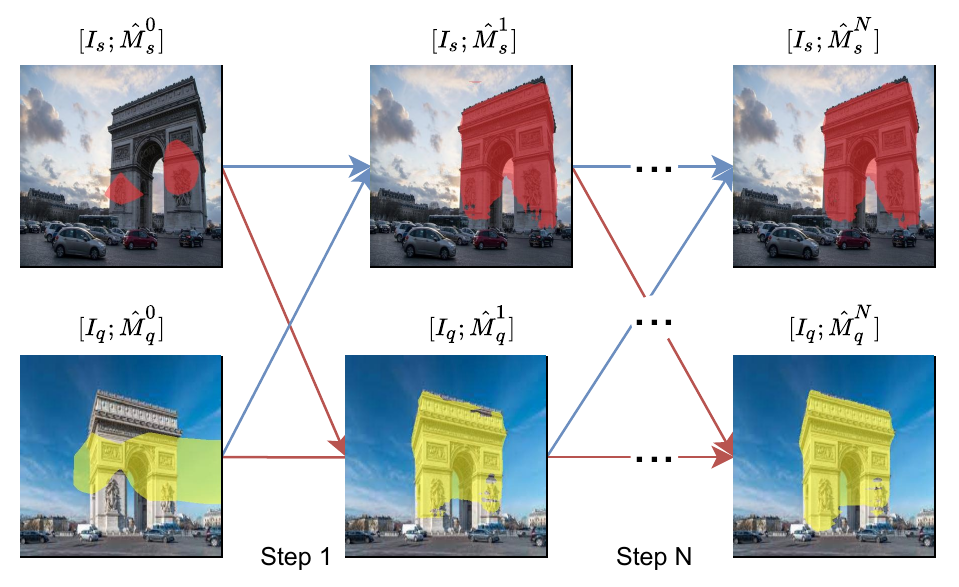}
	\caption{Illustration of the iterative mutual mask refinement. In the step $t$, the posterior query mask estimation $\hat{M}_q^t$ is refined with the guidance of corresponding support mask $\hat{M}_s^{t-1}$ and the prior estimation $\hat{M}_q^{t-1}$. The posterior support mask estimation $\hat{M}_s^t$ is simultaneously updated in a symmetric way. During the refinement, the quality of the object masks is improving gradually.}
	\label{fig:imr}
\end{figure}

\subsection{Iterative Mutual Mask Refinement}
\label{sec:refinement}

With a coarse support mask $\hat{M}_s^0$, an intuitive solution is to feed it into the few-shot segmentation model to obtain a more delicate query mask. Similarly, the finer query mask could then result in the finer support mask,  which is more suitable as the indeed support input. Therefore, an iterative strategy could improve the final prediction gradually by support-query mutual refinement, as illustrated in~\cref{fig:imr}.\par

Nevertheless, the distribution of input masks from different steps varies, which degrades the performance of the model.
Therefore, we propose an \haohan{Iterative Mutual Refinement (IMR) module} to capture and adapt to various input distributions. 

Specifically, we first introduce a hidden state $h$ to reflect how well the input mask is estimated. Thus in the $t$ step, we already have the image feature $F_s, F_q$, the prior estimation of support and query masks $\hat{M}_s^{t-1},\hat{M}_q^{t-1}$, as well as the stored previous state $h_s^{t-1}, h_q^{t-1}$. In order to predict $\hat{M}_q^{t}$, we first encode the information of support images and masks into the query feature, and obtain the correlation volume $C_q^t$, which is consistent with the regular few-shot segmentation. Then $F_q, C_q^t, \hat{M}_q^{t-1}$ are concatenated as $x^t_q$, along with $h_q^{t-1}$ as the input to update the $h_q^{t}$. The whole process of state updating derived from the gated recurrent unit is formulated as: 
\begin{equation}
\begin{aligned}
&z^t_q = \sigma(W_\text{conv1,1} \cdot [x^t_q, h_q^{t-1}]), \\
&r^t_q = \sigma(W_\text{conv1,2} \cdot [x^t_q, h_q^{t-1}]), \\
&\widetilde{h}_q^{t} = \tanh(W_\text{conv2} \cdot [r^t_q \otimes h_q^{t-1}, x^t_q]), \\
&h_q^t = (1-z^t_q) \otimes h_q^{t-1} + z^t_q \otimes \widetilde{h}_q^{t}.\\
\end{aligned}
\end{equation}
where $W_\text{conv1,1}, W_\text{conv1,2}, W_\text{conv2}$ are implemented by $1\times 1$ convolutions followed by $3\times 3$ group convolutions with 16 groups, and $[\cdot], \otimes, \sigma$ denote concatenation, Hadamard product and sigmoid function, respectively. For briefness, we only list how $h_q^{t}$ is updated, and $h_s^{t}$ is updated in the symmetrical way. After that, the updated $h_q^{t}$, as well as the prior estimation $M_q^{t-1}$, are utilized for further prediction. \haohan{Detailed analysis is provided in the supplementary materials.}

\subsection{Model Architecture and Objective}
\label{sec:architecture}

Our main pipeline consists of the \haohan{Iterative Mutual Refinement} module, backbone, neck, and head. The last three modules are abstract from the regular few-shot segmentation methods according to their respective roles, and could be replaced with corresponding modules in different basic methods. \par

In this work we reuse the architecture of HSNet \cite{hsnet} for its excellent performance. Concretely, we adopt ResNet50 \cite{resnet} or VGG16 \cite{vgg} as backbone to extract feature $F_s, F_q$. Then we adopt the well-designed 4D convolutions followed by multi-scale feature fusion as the neck to model the correlation of $F_s$ and $F_q$, and obtain the intermediate output $C_q$. Please refer to \cite{hsnet} for more details. \par

For the mask prediction, we make a minor modification based on the \haohan{original} HSNet head, in order to leverage the prior mask estimation $\hat{M}^{t-1}$, which is formulated as follows:
\begin{gather}
    \tilde{M}^t = \text{Head}_0(C^t +h^{t}), \\
    \hat{M}^t = \text{Softmax} \left(W_\text{h2} \cdot \text{GeLU} (W_\text{h1} \cdot [\tilde{M}^t, \hat{M}^{t-1}])\right).
\end{gather}
where $\text{Head}_0$ is the original head in HSNet, and $W_{h1}, W_{h2}$ is implemented by 1 $\times$ 1 convolutional layers.

\begin{table}[t]
\centering
\small
\resizebox{\columnwidth}{!}{
\centering
\begin{tabular}{@{}cl|c|rrrrr@{}}
\toprule
\multirow{2}{*}{Backbone} & \multirow{2}{*}{Method} & \multirow{2}{*}{A. Type} & \multicolumn{5}{c}{$\text{Pascal}\textit{-}5^i$} \\ \cmidrule(l){4-8} 
 &  &  & $5^{0}$ & $5^{1}$ & $5^{2}$ & $5^{3}$ & Mean \\ \midrule
\multirow{8}{*}{VGG16} & RPMM~\cite{yang2020pmm} & Pixel & 47.1 & 65.8 & 50.6 & 48.5 & 53.0\\
 & ASR~\cite{anti-aliasing-fss} & Pixel & 49.2 & 65.4 & 52.6 & 51.3 & 54.6 \\
 & SS-PANet~\cite{bmvc-fss} & Pixel & 49.3 & 60.8 & 53.9 & 45.2 & 52.3 \\
 & SS-PFENet~\cite{bmvc-fss} & Pixel & 54.5 & \textbf{67.4} & \textbf{63.4} & \textbf{54.0} & \textbf{59.8} \\ 
 & HSNet~\cite{hsnet} & Pixel & \textbf{59.6} & 65.7 & 59.6 & {54.0} & 59.7  \\
 \cmidrule(lr){2-8}
 & PANet~\cite{panet} & Box & -- & -- & -- & -- & 45.1 \\
 & \cite{Lee_2022_WACV} & Image & 36.5 & 51.7 & 45.9 & 35.6 & 42.4 \\ 
 & Ours (IMR-HSNet) & Image & \textbf{58.2} & \textbf{63.9} & \textbf{52.9} & \textbf{51.2} & \textbf{56.5} \\
 \midrule
\multirow{9}{*}{ResNet50} & PFENet~\cite{pfenet} & Pixel & 61.7 & 69.5 & 55.4 & 56.3 & 60.8 \\
 & ASR~\cite{anti-aliasing-fss} & Pixel & 53.8 & 69.6 & 51.6 & 52.8 & 56.9 \\
 & ASGNet~\cite{adaptive-fss} & Pixel & 58.8 & 67.9 & 56.8 & 53.7 & 59.3 \\
 & SCL~\cite{self-guided-fss} & Pixel & 63.0 & 70.0 & 56.5 & 57.7 & 61.8  \\ 
 & HSNet~\cite{hsnet} & Pixel & \textbf{64.3} & \textbf{70.7} & \textbf{60.3} & \textbf{60.5} & \textbf{64.0} \\ \cmidrule(lr){2-8}
 & CANet~\cite{canet} & Box & -- & -- & -- & -- & 52.0 \\
 & \cite{ijcai-wfss}+CANet & Image & 49.5 & 65.5 & 50.0 & 49.2 & 53.5 \\
 & \cite{ijcai-wfss}+PANet & Image & 42.5 & 64.8 & 48.1 & 46.5 & 50.5 \\
 & Ours (IMR-HSNet) & Image & \textbf{62.6} & \textbf{69.1} & \textbf{56.1} & \textbf{56.7} & \textbf{61.1} \\
\bottomrule
\end{tabular}
}
\caption{Comparisons with \whh{regular and weakly supervised few-shot semantic segmentation methods} on $\text{Pascal}\textit{-}5^i$.}
\label{pascal}
\end{table}

In order to accelerate convergence, we add the intermediate supervision to the output mask of each step. As a result, the overall loss function is defined as follows:
\begin{equation}
\mathcal{L}_\text{all} = \sum_{t=1}^N \lambda_t \cdot [\mathcal{L}_\text{seg}(\hat{M}_q^{t}, M_q) + w_s \cdot \mathcal{L}_\text{seg}(\hat{M}_s^{t}, M_s)].
\end{equation}
where $N$ denotes the number of the total steps in the refinement, and $M_s, M_q$ are the ground-truth support and query mask. We adopt cross entropy loss  $\mathcal{L}_\text{seg}$ during training and $\lambda_t$ and $w_s$ are the \haohan{hyperparameters}.\par

\section{Experimental Results}

\subsection{Datasets and Evaluation Metrics}
We follow almost the same settings with regular few-shot segmentation, and the only difference is that the ground-truth masks of support images are replaced by the label text.
we evaluate our framework in two widely-used datasets, \textit{i.e.}, $\text{Pascal}\textit{-}5^i$ \cite{pascal5i}, $\text{COCO}\textit{-}20^i$ \cite{coco20i}, and conduct the cross-validation over all the folds in each dataset. For each fold $i$, samples in the remaining folds serve as the training data, while the 1000 episodes (support-query pairs) are sampled randomly as the testing data. We adopt the mean intersection over union (mIoU) to measure the performance, which averages the IoU of all classes in the testing data to avoid the bias to the specific categories. \par

\subsection{Implementation Details}
We iteratively refine the estimation of support and query masks twice in $\text{Pascal}\textit{-}5^i$, and thrice in $\text{COCO}\textit{-}20^i$ due to the more complex scene. \whh{It is actually a trade-off between effectiveness and efficiency, as we empirically find that the gain of performance gradually converges with the step number increasing.} For a fair comparison, we utilize the ImageNet pretrained VGG16 and ResNet50 as our backbone to extract feature, and the pre-trained CLIP model for initial mask generation. For the CLIP model, the vision and text encoder are a modified ResNet50 and transformer, respectively. The parameters of the backbone and CLIP are fixed during training to avoid the undesirable representation bias towards the base class.

\begin{table}[t]
\small
\resizebox{\columnwidth}{!}{
\centering
\begin{tabular}{@{}cl|c|rrrrr@{}}
\toprule
\multirow{2}{*}{Backbone} & \multirow{2}{*}{Method} & \multirow{2}{*}{A. Type} & \multicolumn{5}{c}{$\text{COCO}\textit{-}20^i$} \\ \cmidrule(l){4-8} 
 &  &  & $20^{0}$ & $20^{1}$ & $20^{2}$ & $20^{3}$ & Mean \\ \midrule
\multirow{7}{*}{VGG16} & PFENet~\cite{pfenet} & Pixel & 35.4 & 38.1 & 36.8 & 34.7 & 36.3\\ \specialrule{0em}{.7pt}{.7pt}
 & SAGNN~\cite{gnn-fss} & Pixel & 35.0 & \textbf{40.5} & \textbf{37.6} & 36.0 & 37.3 \\ \specialrule{0em}{.7pt}{.7pt}
 & SS-PANet~\cite{bmvc-fss} & Pixel & 29.8 & 21.2 & 26.5 & 28.5 & 26.2 \\ \specialrule{0em}{.7pt}{.7pt}
 & SS-PFENet~\cite{bmvc-fss} & Pixel & \textbf{35.6} & 39.2 & \textbf{37.6} & \textbf{37.3} & \textbf{37.5} \\
 \cmidrule(lr){2-8}
 & PANet~\cite{panet} & Box & 12.7 & 8.7 & 5.9 & 4.8 & 8.0  \\ \specialrule{0em}{.7pt}{.7pt}
 & \cite{Lee_2022_WACV} & Image & 24.2 & 12.9 & 17.0 & 14.0 & 17.0 \\ \specialrule{0em}{.7pt}{.7pt}
 & Ours (IMR-HSNet) & Image & \textbf{34.9} & \textbf{38.8} & \textbf{37.0} & \textbf{40.1} & \textbf{37.7}  \\ 
 \midrule
\multirow{9}{*}{ResNet50} & ASR~\cite{anti-aliasing-fss} & Pixel & 29.9 & 35.0 & 31.9 & 33.5 & 32.6  \\
 & RePRI~\cite{ft-cvpr-fss} & Pixel & 31.2 & 38.1 & 33.3 & 33.0 & 34.0 \\
 & ASGNet~\cite{adaptive-fss} & Pixel & -- & -- & -- & -- & 34.6 \\
 & \cite{mining-fss} & Pixel & \textbf{46.8} & 35.3 & 26.2 & 27.1 & 33.9 \\ 
 & \cite{meta-class-fss} & Pixel & 34.9 & 41.0 & 37.2 & 37.0 & 37.5 \\
 & CWT~\cite{ft-iccv-fss} & Pixel & 32.2 & 36.0 & 31.6 & 31.6 & 32.9 \\
 & HSNet~\cite{hsnet} & Pixel & 36.3 & \textbf{43.1} & \textbf{38.7} & \textbf{38.7} & \textbf{39.2} \\ \cmidrule(lr){2-8}
 & \cite{ijcai-wfss} & Image & -- & -- & -- & -- & 15.0 \\
 & Ours (IMR-HSNet) & Image & \textbf{39.5} & \textbf{43.8} & \textbf{42.4} & \textbf{44.1} & \textbf{42.4} \\
\bottomrule
\end{tabular}
}
\caption{Comparisons with \whh{regular and weakly supervised few-shot semantic segmentation methods} on $\text{COCO}\textit{-}20^i$. }
\label{coco}

\end{table}

\haohan{Owing} to our refinement in the second stage, we adopt the most superficial \leone{prompt text} ``{a photo of a} \texttt{[class]}'' for the input to the text encoder, where \texttt{[class]} is replaced by the label of specific categories. 

Consistent with HSNet, we resize the input into 400 $\times$ 400 in both the training and testing stage, and we do not adopt any data augmentation strategies. The network is trained with Adam optimizer with learning rate $2 e^{-4}$, and all \haohan{hyperparameters} $\lambda_t$ and $w_s$ are set to one by default. \whh{However, our method is robust to the choice of these hyperparameters, as shown in the supplementary materials. }\par

\newcommand{\visheight}{0.085\textwidth}
\newcommand{\viswidth}{0.15\textwidth}
\newcommand{\visCsize}{0.33\columnwidth}

\begin{figure*}[t]
	\centering
	\scriptsize
	\setlength\tabcolsep{0.5pt}
	\begin{tabular}{@{}cC{\visCsize}C{\visCsize}C{\visCsize}C{\visCsize} | C{\visCsize}C{\visCsize}@{}}
	    & ``\texttt{Hair Drier}" & ``\texttt{Cell Phone}" &``\texttt{Carrot}" &``\texttt{Motorcycle}" &``\texttt{Dinosaur}" & \tiny{``\texttt{Leaning Tower of Pisa}"} \\
		\rotatebox[origin=l]{90}{Support} & 
		\includegraphics[height=\visheight, width=\viswidth]{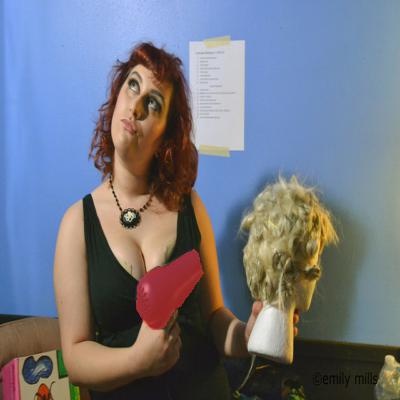} &
		\includegraphics[height=\visheight, width=\viswidth]{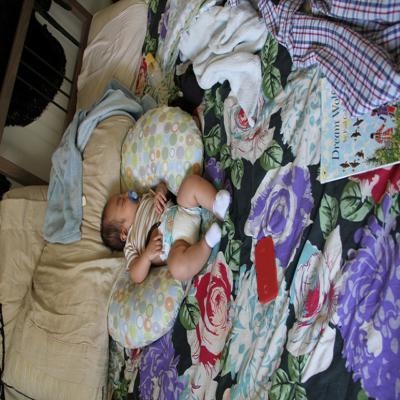} &
		\includegraphics[height=\visheight, width=\viswidth]{pic/qualitative/1s.jpg} &
		\includegraphics[height=\visheight, width=\viswidth]{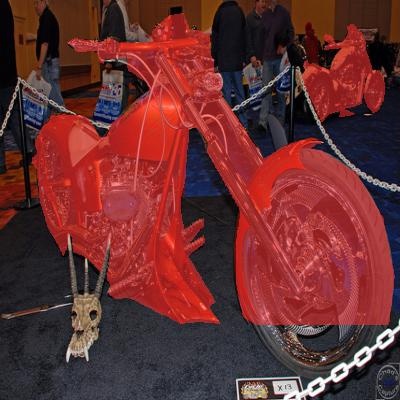} &
		\includegraphics[height=\visheight, width=\viswidth]{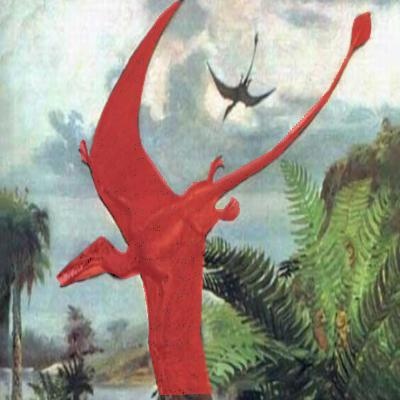} &
		\includegraphics[height=\visheight, width=\viswidth]{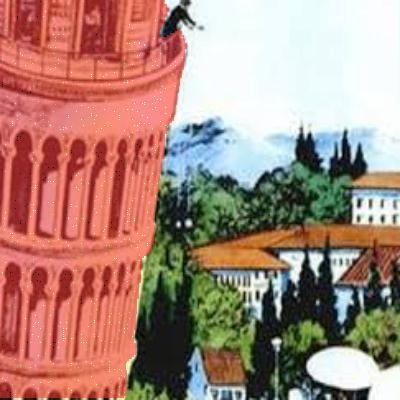} 
		\\
		\rotatebox[origin=l]{90}{Query} & 
		\includegraphics[height=\visheight, width=\viswidth]{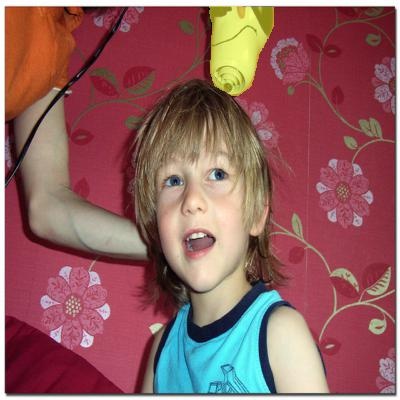} &
		\includegraphics[height=\visheight, width=\viswidth]{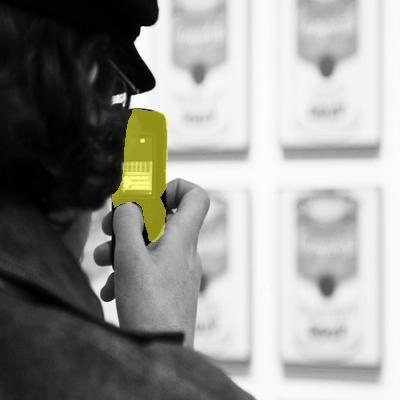} &
		\includegraphics[height=\visheight, width=\viswidth]{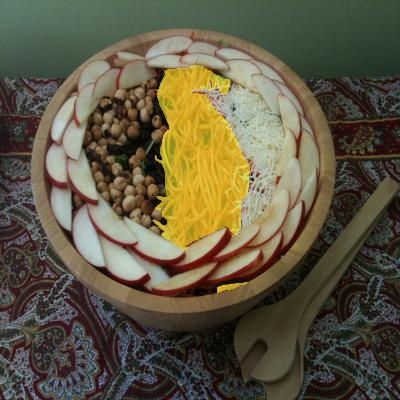} &
		\includegraphics[height=\visheight, width=\viswidth]{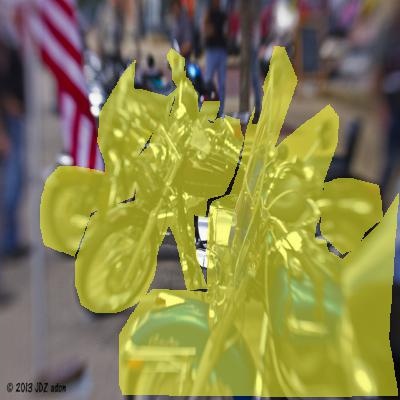} &
		\includegraphics[height=\visheight, width=\viswidth]{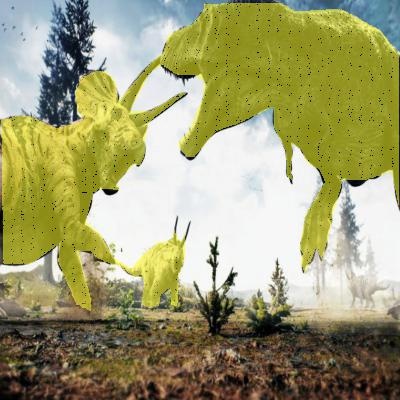} &
		\includegraphics[height=\visheight, width=\viswidth]{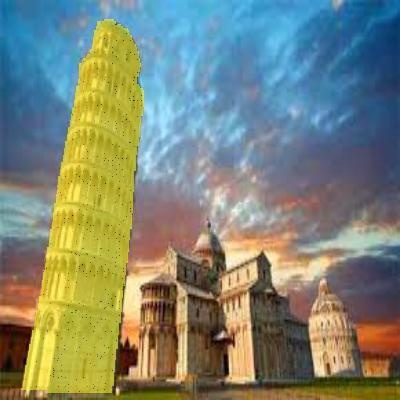} 
		\\
		\rotatebox[origin=l]{90}{HSNet} &
		\includegraphics[height=\visheight, width=\viswidth]{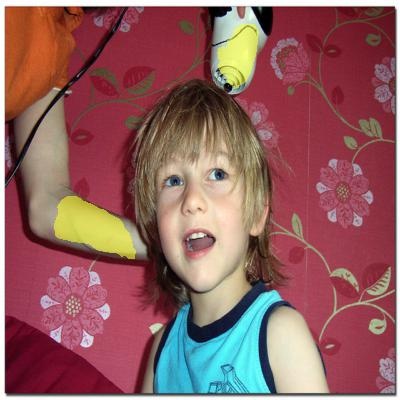} &
		\includegraphics[height=\visheight, width=\viswidth]{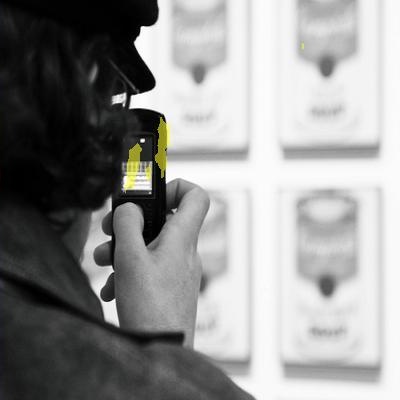} &
		\includegraphics[height=\visheight, width=\viswidth]{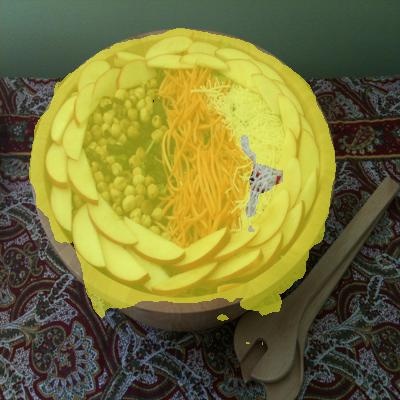} &
		\includegraphics[height=\visheight, width=\viswidth]{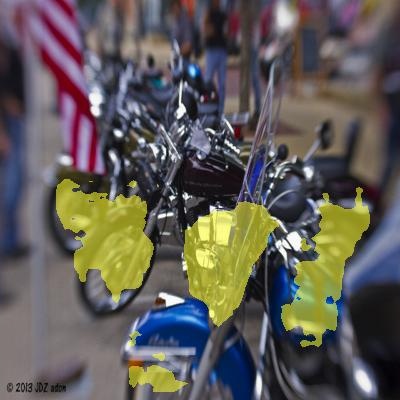} &
		\includegraphics[height=\visheight, width=\viswidth]{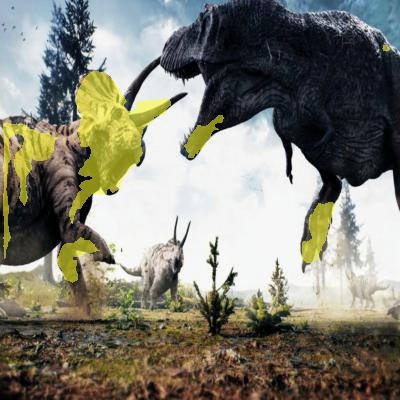} &
		\includegraphics[height=\visheight, width=\viswidth]{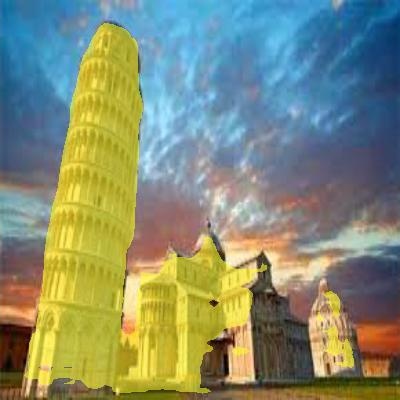} 
		\\
		\rotatebox[origin=l]{90}{Ours} & 
		\includegraphics[height=\visheight, width=\viswidth]{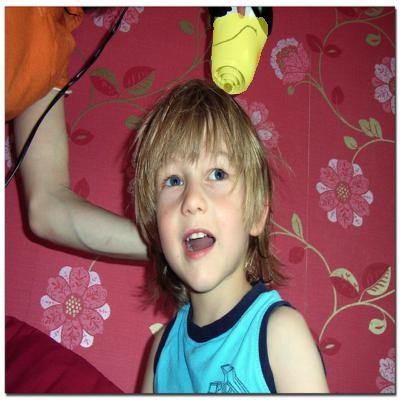} &
		\includegraphics[height=\visheight, width=\viswidth]{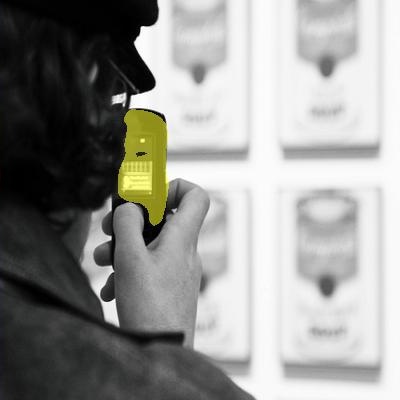} &
		\includegraphics[height=\visheight, width=\viswidth]{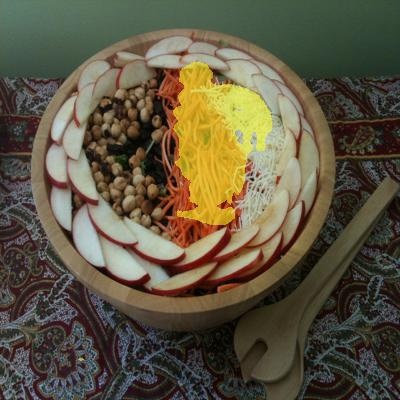} &
		\includegraphics[height=\visheight, width=\viswidth]{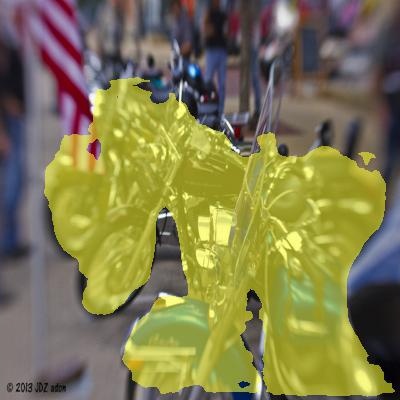} &
		\includegraphics[height=\visheight, width=\viswidth]{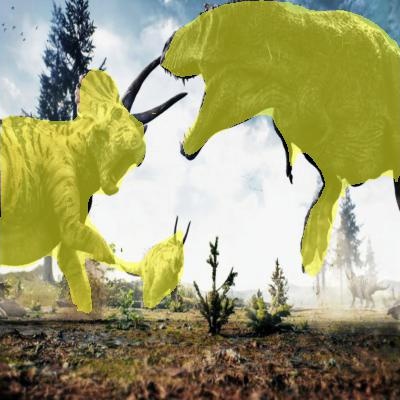} &
		\includegraphics[height=\visheight, width=\viswidth]{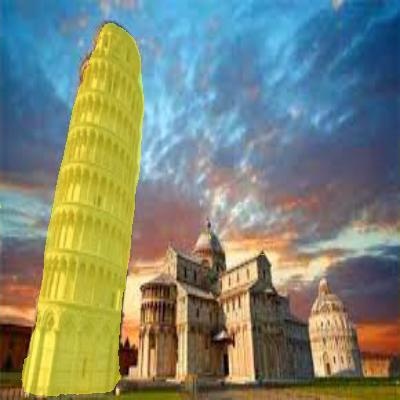} 
		\\
	\end{tabular}

	\caption{Qualitative visualization comparing with the regular few-shot segmentation method HSNet~\protect\cite{hsnet}. The first four columns are from the $\text{COCO}\textit{-}20^i$, and the last two columns are from web to show the performance in-the-wild.}
	\label{fig:visualization}
\end{figure*}

\subsection{Comparison with State-of-the-art}

In this subsection, we compare our method with the state-of-the-art weakly supervised and regular few-shot segmentation models in the 1-shot setting. ~\cref{pascal} presents the results in the ${\rm Pascal} \textit{-}5^i$. It could be seen that our method sets new state of the arts with both VGG16 and ResNet50 backbone in the weakly supervised setting, and outperforms its counterparts with either box-level or image-level annotation, by a large margin (11\% with VGG16 and \haohan{7.6\%} with ResNet50). Furthermore, benefiting from our iterative strategy, our weakly supervised method surpasses most regular few-shot models.

~\cref{coco} summarizes the results in the more challenge benchmark $\text{COCO}\textit{-}20^i$. When more objects with various scales appear in the query image, the performance of other weakly supervised models falls much more behind that of fully supervised models. However, our method turns out to be robust to this complex scene, and thus accordingly, the superiority of our method is more significant. Specifically, we achieve more than larger performance gains (20\% and 27\% for VGG16 and ResNet) compared to the state-of-the-art ones using weak annotation. Furthermore, adopting ResNet50 as the backbone, our method in the weakly supervised setting achieves \haohan{42.4 
mIoU}, which even outperforms the state-of-the-art model in the standard setting.

\begin{table}[t]
\centering
\resizebox{0.97\columnwidth}{!}{
\begin{tabular}{@{}lcc|ccccc@{}}
\toprule
Method & Shot & Backbone & $20^{0}$ & $20^{1}$ & $20^{2}$ & $20^{3}$ & Mean \\ \midrule
ZS3Net~\cite{bucher2019zero} & Zero & Res101 & 18.8 & 20.1 & 24.8 & 20.5 & 21.1 \\
L-Seg~\cite{anonymous2022languagedriven} & Zero & Res101 & 22.1 & 25.1 & 24.9 & 21.5 & 23.4 \\
L-Seg~\cite{anonymous2022languagedriven} & Zero & ViT-L/16 & 28.1 & 27.5 & 30.0 & 23.2 & 27.2 \\
Ours & Zero & Res50 & \textbf{37.9} & \textbf{41.8} & \textbf{41.3} & \textbf{43.3} & \textbf{41.1} \\ \midrule
Ours & One & Res50 & 39.5 & 43.8 & 42.4 & 44.1 & 42.4 \\
Ours & Five & Res50 & \textbf{40.7} & \textbf{46.0} & \textbf{45.0} & \textbf{46.0} & \textbf{44.4} \\ \bottomrule
\end{tabular}}
\caption{Comparisons with \whh{zero-shot semantic segmentation} methods on $\text{COCO}\textit{-}20^i$.}
\label{zeroshot}
\end{table}

\begin{table}[t]
\centering
\small
\resizebox{0.8\columnwidth}{!}{
\centering
\begin{tabular}{@{}lc|ccccc@{}}
\toprule
Methods & Shot & $5^{0}$ & $5^{1}$ & $5^{2}$ & $5^{3}$ & Mean \\ \midrule
Cosine Sim & Zero & 21.4 & {39.9} & 27.9 & {32.1} & {30.3} \\  \specialrule{0em}{1pt}{1pt}
CLIP-CAM & Zero & {24.3} & 27.6 & {29.7} & 29.0 & 27.6 \\ \specialrule{0em}{1pt}{1pt}
Cosine Sim  & One & 38.9 & 56.2 & 39.1 & 39.5 & 43.4 \\  \specialrule{0em}{1pt}{1pt}
CLIP-CAM  & One & {50.2} & {58.6} & {49.0} & {47.5} & {51.3} \\ \bottomrule
\end{tabular}
}
\caption{Comparisons of initial mask generation methods.}
\label{initialmask}
\end{table}

\begin{table}[t]
\centering
\small
\resizebox{0.97\columnwidth}{!}{
\begin{tabular}{@{}l|ccccc@{}}
\toprule
Methods & $5^{0}$ & $5^{1}$ & $5^{2}$ & $5^{3}$ & Mean \\ \midrule
+HSNet (pretrained) & 50.2 & 58.6 & 49.0 & 47.5 & 51.3 \\
+HSNet (mutual) & 59.5 & 67.5 & 54.1 & 53.8 & 58.7\\
+HSNet (retrained) & 60.9 & 65.8 & 53.1 & 54.7 & 58.6  \\
+HSNet (retrained+mutual) & 56.2 & 63.8 & 47.0 & 49.6 & 54.2  \\

Ours (IMR-HSNet) & 62.6 & 69.1 & 56.1 & 56.7 & 61.1 \\ \bottomrule
\end{tabular}
}
\caption{Comparisons of mask refinement methods.}
\label{refine}
\end{table}

\begin{table}[t]
\centering
\resizebox{0.8\columnwidth}{!}{
\begin{tabular}{@{}l|ccccc@{}}
\toprule
Methods  & $5^{0}$ & $5^{1}$ & $5^{2}$ & $5^{3}$ & Mean \\ \midrule
CLIP-CAM & 24.3 & 27.6 & 29.7 & 29.0 & 27.6 \\  \specialrule{0em}{1pt}{1pt}
CLIP-CAM / PFENet &  44.2 & 51.1 & 49.0 & 47.9 & 48.0 \\  \specialrule{0em}{.7pt}{.7pt}
CLIP-CAM / HSNet &  50.2 & 58.6 & 49.0 & 47.5 & 51.3 \\  \specialrule{0em}{.7pt}{.7pt}
Ours (IMR-PFENet) &  53.4 & 53.7 & 54.2 & 53.3 & 53.7 \\  \specialrule{0em}{.7pt}{.7pt}
Ours (IMR-HSNet) &  62.6 & 69.1 & 56.1 & 56.7 & 61.1 \\ \specialrule{0em}{.7pt}{.7pt} \bottomrule
\end{tabular}
}
\caption{Generalization ability with other few-shot segmentation architectures.}
\label{pfenet}

\end{table}

In order to show the comparison more intuitively, we visualize the results of some of the \haohan{challenging} query images, which are segmented by state-of-the-art HSNet (standard few-shot setting) and our method (weakly supervised setting). As shown in~\cref{fig:visualization}, our method could perform better in segmenting tiny objects (1st and 2nd columns), semantically confusing objects (3rd column), multiple objects (4th column), and objects in the wild (last two columns). \whh{This is probably because our method is able to associate support and query objects with a large difference in appearance with the help of label text, and more likely to predict the complete masks with complex shapes thanks to the prior mask estimation in each step of the refinement. }\par

\subsection{Comparison with Zero-shot Segmentation}
Weakly supervised few-shot segmentation requires support label text rather than the support mask to segment the query image. From this perspective, it has a lot in common with zero-shot semantic segmentation (ZSS), where only the label text without the support image is available. 
Therefore, it is natural to compare our method with zero-shot counterparts. 
However, the comparison is non-trivial due to the inconsistent testing protocol between zero-shot and few-shot segmentation as well as the lack of zero-shot datasets and baselines. Thus, 
we still adopt the few-shot benchmark, \textit{i.e.}, $\text{Pascal}\textit{-}5^i$ and $\text{COCO}\textit{-}20^i$ for evaluation.

In order to apply our framework to the zero-shot segmentation, we need to make some modifications. \whh{Therefore,} we take the query image as the input of where the support image is required. \whh{Concretely, we regard $\{I_q, \hat{M}_q^{t-1}\}$ as the $\{I_s, \hat{M}_s^{t-1}\}$ to predict $\hat{M}_q^{t}$, and $\hat{M}_s^t$ is equal to $\hat{M}_q^t$, then repeat the refinement to obtain final $\hat{M}_q^N$.} In this way, the query image is segmented with the guidance by itself. We compare our method with competitive zero-shot methods, \textit{e.g.} ZS3Net~\cite{bucher2019zero}, and L-Seg~\cite{anonymous2022languagedriven}.\par

As reported in~\cref{zeroshot}, our method achieves superior performance in the zero-shot setting with a mean mIoU score of \haohan{41.1}. Without being designed \haohan{and trained} specifically for zero-shot segmentation and strong backbone ResNet101, our method could outperform the recent zero-shot approaches \haohan{by a significant margin}. However, the zero-shot performance still falls behind that of the few-shot version, which proves the importance of the support image, even without the precise object mask. \par

\subsection{Ablation Study}

We conduct the ablation study to investigate the effects of components in our framework. We adopt the ResNet50 as our backbone, and evaluate in the ${\rm Pascal} \textit{-}5^i$ dataset.

\paragraph{Initial Mask Generation.} 
In~\cref{sec:mask_generation}, we introduce two kinds of ways to generate the initial support and query mask $\hat{M}_s^0, \hat{M}_q^0$, \whh{\textit{i.e.}, the pixel-level cosine similarity between text and dense visual feature, and the Grad CAM of similarity between text and global visual feature.} We further compare the two strategies in the zero-shot and few-shot settings. For zero-shot segmentation, we adopt the generated query mask $\hat{M}_q^0$ as the predicted query mask. And for few-shot segmentation, we feed the generated support mask $\hat{M}_s^0$ into the pre-trained HSNet as the ground-truth support mask to segment the query images. 

~\cref{initialmask} reports the results of the comparison. From the first two rows (zero-shot setting), it can be seen that the quality of $\hat{M}^0$ is similar. Nevertheless, when using the generated mask as the support mask, the results of Grad CAM show clear superiority with nearly 10\%. The reason could be that the Grad CAM better leverages the architecture and weights of the pre-trained vision encoder, and thus reflects the strong generality. As a result, we utilize this strategy to generate the initial support and query mask estimation.\par

\paragraph{Iterative Mutual Refinement.} 
We further compare different iterative strategies in~\cref{refine} to evaluate the effectiveness of IMR. We set the method without \haohan{Iterative Mutual Refinement} as the baseline, which is implemented by feeding $\hat{M}_s^0$ to the pretrained HSNet, denoted as \textit{+HSNet (pretrained)}. We start with evaluating a simple iterative strategy, where we first predict $\hat{M}_q^1$ with $\hat{M}_s^0$, then update $\hat{M}_s^2$ with $\hat{M}_q^1$, and get the final prediction $\hat{M}_q^3$ guided by $\hat{M}_s^2$, denoted as \textit{+HSNet (mutual)}. Simultaneously, in order to adapt to the condition where the clean ground-truth support mask is unavailable, we trained HSNet with coarse support masks generated by \haohan{CLIP-CAM}. Similarly, we feed $\hat{M}_s^0$ to the retrained HSNet (\textit{+HSNet (retrained)}), and refine the estimation of support and query mask iteratively (\textit{+HSNet (retrained + mutual)}).\par

As shown in~\cref{refine}, both iterative strategies outperform the simple baseline.
Therefore, iterative refining mask estimation may be the key to weakly supervised few-shot segmentation. Moreover, our method with IMR still shows great superiority due to the architecture designed with hidden states to fit the varying input distribution during the refinement. 

\paragraph{Generalization of Our Method.}
As mentioned above, our framework is plug-and-play and model-agnostic. In order to verify this, we evaluate our framework with another few-shot segmentation model PFENet \cite{pfenet}. Concretely, we adopt the last convolutional layers of PFENet as the \textit{head} in our method, and the other modules, including the prior generation and feature enrichment module (FEM), as the \textit{neck}. We keep all the \haohan{hyperparameters} the same with the origin implementation of PFENet. We compare our method built on PFENet with corresponding baselines, where we feed the $\hat{M}_s^0$ as the support mask to the pre-trained PFENet. 

As shown in~\cref{pfenet}, with the well-designed iterative strategy, our method outperforms the baseline by around 6\%. Hence, it could be concluded that our method is robust to the selection of the few-shot segmentation model.

\section{Conclusion}

In this work, we present a general framework to tackle the weakly supervised few-shot segmentation, which contains coarse masks generation from label text and \haohan{iteratively} and mutually refine support and query mask. 
Extensive experiments show that our method outperforms the state-of-the-art counterparts by a large margin. 
One limitation is that although our method is able to refine segmentation from coarse masks, it is unclear whether finer initial masks from prompt engineering or post-process such as dense CRF could help performance.We leave it for future work. \par
\section*{Acknowledgements}
This work is partially supported by the NSFC fund (61831014), in part by the Shenzhen Science and Technology Project under Grant (ZDYBH201900000002, JCYJ20180508152042002, CJGJZD20200617102601004) \par

\bibliographystyle{named}
\bibliography{ijcai22}

\end{document}


\newcolumntype{C}[1]{>{\centering\arraybackslash}p{#1}} %
\captionsetup[subfigure]{subrefformat=simple,labelformat=simple}
\renewcommand\thesubfigure{(\alph{subfigure})}

\newcommand{\leone}[1]{\textcolor{black}{#1}}
\newcommand{\haohan}[1]{\textcolor{black}{#1}}

\maketitle
\haohan{Thanks for reading the supplementary materials. Here we will introduce more details of our whole pipeline including the initial mask generation and some basic modules from HSNet \cite{hsnet}. We also show experiment results on the extension to five-shot and zero-shot segmentation and more ablation study, as well as plentiful qualitative comparisons.}\par

\section{Architecture Details}
\subsection{Initial Mask Generation}
In this subsection we show how we generate initial support mask by pixel-level cosine similarity or Grad CAM in details. \par
Generating initial mask directly from cosine similarity requires the dense query feature with spatial structure maintained. However, since CLIP adopts an attention-based learnable pooling strategy $p(\cdot)$ to down-sample the last convolutional feature $f_{conv}\in \mathbb{R}^{h \times w \times 2048}$ into  $f_{v}\in \mathbb{R}^{1 \times 1 \times 1024}$, it is impractical to compute the dense cosine similarity between $f_{conv}$ and the text feature $f_{t}\in \mathbb{R}^{1 \times 1 \times 1024}$. Thus following \cite{denseclip-zero-shot}, we make some modifications to $p(\cdot)$ to unify the different dimensions of $f_{conv}$ and $f_{t}$.  Specifically, $p(\cdot)$ consists of three linear layers $q(\cdot), k(\cdot), v(\cdot),$ to project the flattened $f_{conv}\in \mathbb{R}^{hw \times 2048}$ for the self-attention module, and the final linear layer $c(\cdot)$ to process the output of self-attention module to obtain $f_v$. We map the linear layers $v(\cdot)$ and $c(\cdot)$ into two $1 \times 1$ convolutions, which is then directly applied on $f_{conv}$ to obtain $\widetilde{f}_{conv}\in \mathbb{R}^{h \times w \times 1024}$. Therefore, we could then compute the cosine similarity between $f_t$ and $\widetilde{f}_{conv}$ to get the similarity map $S$. Furthermore, for each position $(x,y)$, we normalize the $S(x,y)$ by:
\begin{equation}
\begin{aligned}
\widetilde{S}(x,y) = \frac{S(x,y)-\min{S(x,y)}}{\max{S(x,y)}-\min{S(x,y)} + \sigma}.
\end{aligned}
\end{equation}
where we set $\sigma=0.0001$ to avoid a denominator of zero. Finally we binarize $\widetilde{S}$ with the threshold 0.5 to obtain the $\hat{M}^0$.\par

Generating initial mask from Grad CAM is easier to implement. We compute Grad CAM \haohan{heatmap} $G$ of the output of the last bottleneck of the vision encoder, according to the cosine similarity between $f_v$ and $f_t$. We scale $G$ to the size of input image, and binarize $G$ with the threshold 0.5 to obtain the $\hat{M}^0$.\par

Besides, it is worth mentioning that in order to ensure differentiability, we adopt the soft mask without binarization during training with IMR.\par

\subsection{HSNet}
In this subsection we introduce more details of HSNet \cite{hsnet}, as we reuse many of its modules in our work. The origin architecture of HSNet is shown in \cref{hsnet}, where we divide all the modules into backbone, neck and head, and we omit the backbone for simplification. When adopting ResNet50 as the backbone, HSNet takes all feature maps from each bottleneck in \texttt{conv3\_x, conv4\_x, conv5\_x} as the input, which are denoted as $F^{(3)}, F^{(4)}, F^{(5)}$ in \cref{hsnet}. With a well-designed center-pivot 4D convolution and multi-scale feature fusion, HSNet is able to encode the information of support images into query feature to obtain $C_q$, which is further used for predicting query mask. For the specific parameters of each modules such as keneral size and stride, please refer to \cite{hsnet}.\par

\begin{table}[h]
\centering
\resizebox{0.95\columnwidth}{!}{
\begin{tabular}{|c|ccccccccc|}
\hline
\multirow{4}{*}{Input} & \multicolumn{3}{c|}{Branch 1} & \multicolumn{3}{c|}{Branch 2} & \multicolumn{3}{c|}{Branch 3} \\ \cline{2-10} 
 & \multicolumn{1}{c|}{$F_q^{(5)}$} & \multicolumn{1}{c|}{$F_s^{(5)}$} & \multicolumn{1}{c|}{$M_s$} & \multicolumn{1}{c|}{$F_q^{(4)}$} & \multicolumn{1}{c|}{$F_s^{(4)}$} & \multicolumn{1}{c|}{$M_s$} & \multicolumn{1}{c|}{$F_q^{(3)}$} & \multicolumn{1}{c|}{$F_s^{(3)}$} & $M_s$ \\ \cline{2-10} 
 & \multicolumn{3}{c|}{Cosine Sim} & \multicolumn{3}{c|}{Cosine Sim} & \multicolumn{3}{c|}{Cosine Sim} \\ \cline{2-10} 
 & \multicolumn{3}{c|}{$C^{(5)}$} & \multicolumn{3}{c|}{$C^{(4)}$} & \multicolumn{3}{c|}{$C^{(3)}$} \\ \hline
\multirow{16}{*}{Neck} & \multicolumn{3}{c|}{4D Conv} & \multicolumn{3}{c|}{4D Conv} & \multicolumn{3}{c|}{4D Conv} \\ \cline{2-10} 
 & \multicolumn{3}{c|}{GN} & \multicolumn{3}{c|}{GN} & \multicolumn{3}{c|}{GN} \\ \cline{2-10} 
 & \multicolumn{3}{c|}{ReLU} & \multicolumn{3}{c|}{ReLU} & \multicolumn{3}{c|}{ReLU} \\ \cline{2-10} 
 & \multicolumn{3}{c|}{4D Conv} & \multicolumn{3}{c|}{4D Conv} & \multicolumn{3}{c|}{4D Conv} \\ \cline{2-10} 
 & \multicolumn{3}{c|}{GN} & \multicolumn{3}{c|}{GN} & \multicolumn{3}{c|}{GN} \\ \cline{2-10} 
 & \multicolumn{3}{c|}{ReLU} & \multicolumn{3}{c|}{ReLU} & \multicolumn{3}{c|}{ReLU} \\ \cline{2-10} 
 & \multicolumn{3}{c|}{4D Conv} & \multicolumn{3}{c|}{4D Conv} & \multicolumn{3}{c|}{4D Conv} \\ \cline{2-10} 
 & \multicolumn{3}{c|}{GN} & \multicolumn{3}{c|}{GN} & \multicolumn{3}{c|}{GN} \\ \cline{2-10} 
 & \multicolumn{3}{c|}{ReLU} & \multicolumn{3}{c|}{\multirow{2}{*}{ReLU}} & \multicolumn{3}{c|}{\multirow{5}{*}{ReLU}} \\ \cline{2-4}
 & \multicolumn{3}{c|}{Upsampling} & \multicolumn{3}{c|}{} & \multicolumn{3}{c|}{} \\ \cline{2-7}
 & \multicolumn{6}{c|}{Add} & \multicolumn{3}{c|}{} \\ \cline{2-7}
 & \multicolumn{6}{c|}{4D Conv} & \multicolumn{3}{c|}{} \\ \cline{2-7}
 & \multicolumn{6}{c|}{Upsampling} & \multicolumn{3}{c|}{} \\ \cline{2-10} 
 & \multicolumn{9}{c|}{Add} \\ \cline{2-10} 
 & \multicolumn{9}{c|}{4D Conv} \\ \cline{2-10} 
 & \multicolumn{9}{c|}{Pooling} \\ \hline
Correlation & \multicolumn{9}{c|}{$C_q\in \mathbb{R}^{50\times 50\times 128}$} \\ \hline
\multirow{9}{*}{Head} & \multicolumn{9}{c|}{2D Conv} \\ \cline{2-10} 
 & \multicolumn{9}{c|}{ReLU} \\ \cline{2-10} 
 & \multicolumn{9}{c|}{2D Conv} \\ \cline{2-10} 
 & \multicolumn{9}{c|}{ReLU} \\ \cline{2-10} 
 & \multicolumn{9}{c|}{Upsampling} \\ \cline{2-10} 
 & \multicolumn{9}{c|}{2D Conv} \\ \cline{2-10} 
 & \multicolumn{9}{c|}{ReLU} \\ \cline{2-10} 
 & \multicolumn{9}{c|}{2D Conv} \\ \cline{2-10} 
 & \multicolumn{9}{c|}{Upsampling} \\ \hline
\multicolumn{1}{|c|}{Prediction} & \multicolumn{9}{c|}{Mask} \\ \hline
\end{tabular}}
\vspace{-0.1cm}
\caption{\leone{The original architecture of HSNet, abstracted as Backbone (omitted in the table), Neck and Head.} }
\vspace{-0.4cm}
\label{hsnet}
\end{table}

\begin{table}[t]
\resizebox{\columnwidth}{!}{
\begin{tabular}{@{}cl|c|ccccc@{}}
\toprule
\multirow{2}{*}{Backbone} & \multirow{2}{*}{Method} & \multirow{2}{*}{A. Type} & \multicolumn{5}{c}{$\text{Pascal}\textit{-}5^i$} \\ \cmidrule(l){4-8} 
 &  &  & $5^{0}$ & $5^{1}$ & $5^{2}$ & $5^{3}$ & Mean \\ \midrule
\multirow{8}{*}{VGG16} & \multicolumn{1}{l|}{RPMM ECCV20} & Pixel & 50.0 & 66.5 & 51.9 & 47.6 & 54.0 \\
 & \multicolumn{1}{l|}{ASR~\cite{anti-aliasing-fss}} & Pixel & 52.5 & 66.5 & 55.0 & 53.9 & 57.0 \\
 & \multicolumn{1}{l|}{SS-PANet~\cite{bmvc-fss}} & Pixel & 54.5 & 64.5 & 62.3 & 52.0 & 58.3 \\
 & \multicolumn{1}{l|}{SS-PFENet~\cite{bmvc-fss}} & Pixel & 56.9 & \textbf{70.0} & \textbf{68.3} & \textbf{62.1} & \textbf{64.3} \\
 & \multicolumn{1}{l|}{HSNet~\cite{hsnet}} & Pixel & \textbf{64.9} & 69.0 & 64.1 & 58.6 & 64.1 \\ \cmidrule(l){2-8} 
 & \multicolumn{1}{l|}{PANet~\cite{panet}} & Box & -- & -- & -- & -- & 52.8 \\
 & \multicolumn{1}{l|}{\cite{Lee_2022_WACV}} & Image & -- & -- & -- & -- & 45.5 \\
 & \multicolumn{1}{l|}{Ours (IMR-HSNet)} & Image & \textbf{60.5} & \textbf{65.3} & \textbf{55.0} & \textbf{51.7} & \textbf{58.1} \\ \midrule
\multirow{9}{*}{ResNet50} & \multicolumn{1}{l|}{PFENet PAMI20} & Pixel & 63.1 & 70.7 & 55.8 & 57.9 & 61.9 \\
 & \multicolumn{1}{l|}{ASR~\cite{anti-aliasing-fss}} & Pixel & 56.2 & 70.6 & 53.9 & 53.4 & 58.5 \\
 & \multicolumn{1}{l|}{ASGNet~\cite{adaptive-fss}} & Pixel & 63.7 & 70.6 & 64.2 & 57.4 & 63.9 \\
 & \multicolumn{1}{l|}{SCL~\cite{self-guided-fss}} & Pixel & 64.5 & 70.9 & 57.3 & 58.7 & 62.9 \\
 & \multicolumn{1}{l|}{CWT~\cite{ft-iccv-fss}} & Pixel & 61.3 & 68.5 & \textbf{68.5} & 56.6 & 63.7 \\
 & \multicolumn{1}{l|}{HSNet~\cite{hsnet}} & Pixel & \textbf{70.3} & \textbf{73.2} & 67.4 & \textbf{67.1} & \textbf{69.5} \\ \cmidrule(l){2-8} 
 & \multicolumn{1}{l|}{\cite{ijcai-wfss}} & Image & 45.9 & 65.7 & 48.6 & 46.6 & 51.7 \\
 & \multicolumn{1}{l|}{Ours (IMR-HSNet)} & Image & \textbf{63.6} & \textbf{69.6} & \textbf{56.3} & \textbf{57.4} & \textbf{61.8} \\ \bottomrule
\end{tabular}
}
\caption{Comparisons with WFSS and FSS Methods on $\text{Pascal}\textit{-}5^i$ in five-shot setting.}
\label{pascal}
\end{table}

\begin{table}[ht]
\resizebox{\columnwidth}{!}{
\begin{tabular}{@{}cl|c|ccccc@{}}
\toprule
\multirow{2}{*}{Backbone} & \multirow{2}{*}{Method} & \multicolumn{1}{l|}{\multirow{2}{*}{A.Type}} & \multicolumn{5}{c}{$\text{COCO}\textit{-}20^i$} \\ \cmidrule(l){4-8} 
 &  & \multicolumn{1}{l|}{} & $20^{0}$ & $20^{1}$ & $20^{2}$ & $20^{3}$ & Mean \\ \midrule
\multirow{7}{*}{VGG16} & PFENet~\cite{pfenet} & Pixel & 38.2 & 42.5 & \textbf{41.8} & 38.9 & 40.4 \\ \specialrule{0em}{.7pt}{.7pt}
 & SAGNN~\cite{gnn-fss} & Pixel & 37.2 & 45.2 & 40.4 & 40.0 & 40.7 \\ 
 & SS-PANet~\cite{bmvc-fss} & Pixel & 36.7 & 41.0 & 37.6 & 35.6 & 37.7 \\ 
 & SS-PFENet~\cite{bmvc-fss} & Pixel & \textbf{40.4} & \textbf{45.8} & 40.3 & \textbf{40.7} & \textbf{41.8} \\ \cmidrule(l){2-8} 
 & PANet~\cite{panet} & Box & -- & -- & -- & -- & 13.9 \\ 
 & \cite{Lee_2022_WACV} & Image & -- & -- & -- & -- & 17.5 \\ 
 & Ours (IMR-HSNet) & Image & \textbf{34.8} & \textbf{41.0} & \textbf{37.2} & \textbf{39.7} & \textbf{38.2} \\ \midrule
\multirow{9}{*}{ResNet50} & ASR~\cite{anti-aliasing-fss} & Pixel & 31.3 & 37.9 & 33.5 & 35.2 & 34.4 \\
 & RePRI~\cite{ft-cvpr-fss} & Pixel & 38.5 & 46.2 & 40.0 & 43.6 & 42.1 \\
 & ASGNet~\cite{adaptive-fss} & Pixel & -- & -- & -- & -- & 42.5 \\
 &  \cite{mining-fss} & Pixel & \textbf{54.1} & 41.2 & 34.1 & 33.1 & 40.6 \\
 & \cite{meta-class-fss} & Pixel & 37.0 & 40.3 & 39.3 & 36.0 & 38.2 \\
 & CWT~\cite{ft-iccv-fss} & Pixel & 40.1 & 43.8 & 39 & 42.4 & 41.3 \\
 & HSNet~\cite{hsnet} & Pixel & 43.3 & \textbf{51.3} & \textbf{48.2} & \textbf{45.0} & \textbf{46.9} \\ \cmidrule(l){2-8} 
 & \cite{ijcai-wfss}  & Image & -- & -- & -- & -- & 15.6 \\
 & Ours (IMR-HSNet) & Image & \textbf{40.7} & \textbf{46.0} & \textbf{45.0} & \textbf{46.0} & \textbf{44.4} \\ \bottomrule
\end{tabular}
}
\caption{Comparisons with WFSS and FSS Methods on $\text{COCO}\textit{-}20^i$ in five-shot setting.}
\label{coco}
\end{table}

\section{Additional Experiments}

\subsection{Weakly Supervised Five-shot Segmentation}

We follow \cite{hsnet} to inference with $K$ support images. Concretely, we first conduct one-shot segmentation for $K$ times to obtain $K$ binary query mask $\hat{M}_q^{(k)}$. Then for each position $(x,y)$, we aggregate $K$ predicted to get the finial prediction $\hat{M}_q$ by:
\begin{equation}
\begin{aligned}
\hat{M}_{sum} (x,y) = \sum_{k=1}^K \hat{M}_q^k (x,y), \\
M = \max{(\hat{M}_{sum} (x,y))}, \\
\hat{M}_q (x,y) = \frac{\hat{M}_{sum} (x,y)}{M}. \\
\end{aligned}
\end{equation}

Consistent with the standard few-shot segmentation, we conduct the experiments in weakly supervised five-shot setting, and compare our method with the state-of-the-art few-shot segmentation models in both weakly supervised and standard settings.\par

As shown in \cref{pascal} and \cref{coco}, our method keeps the great superiority over other weakly supervised counterparts in both benchmarks. In $\text{Pascal}\textit{-}5^i$, our method surpasses the state-of-the-art weakly supervised models by around 5$\sim$10\%. And in the more challenging $\text{COCO}\textit{-}20^i$, our method achieves larger performance gains of over 20\%. The conclusion from the five-shot experiments stay consistent with that in one-shot setting, which demonstrates the effectiveness of our method, especially in the complex scene.\par

However, compared with results in one-shot setting, the performance gap between our method (weakly supervised) and the state-of-the-art regular few-shot models is a bit bigger. It is acceptable as we adopt a very simple strategy to average five predicted query mask in one-shot way without considering the relation between the support images. Better aggregating the information from multiple support images would benefit the prediction, and we leave it for future work.\par

\subsection{Zero-shot Segmentation on $\text{Pascal}\textit{-}5^i$}
\haohan{In this subsection we report the zero-shot segmentation performance on $\text{Pascal}\textit{-}5^i$. As shown in \cref{zeroshot}, our method outperforms the leading zero-shot segmentation models by around 7\%. It is also worth mentioning that we only adopt the model trained in few-shot setting to tackle the zero-shot segmentation, and more specific designs could further boost our performance. }\par

\begin{table}[t]
\centering
\resizebox{0.97\columnwidth}{!}{
\begin{tabular}{@{}lcc|ccccc@{}}
\toprule
Method & Shot & Backbone & \multicolumn{1}{c}{$5^{0}$} & \multicolumn{1}{c}{$5^{1}$} & \multicolumn{1}{c}{$5^{2}$} & \multicolumn{1}{c}{$5^{3}$} & Mean \\ \midrule
SPNet~\cite{xian2019semantic} & Zero & Res101 & 23.8 & 17.0 & 14.1 & 18.3 & 18.3 \\
ZS3Net~\cite{bucher2019zero} & Zero & Res101 & 40.8 & 39.4 & 39.3 & 33.6 & 38.3 \\
L-Seg~\cite{anonymous2022languagedriven} & Zero & Res101 & 52.8 & 53.8 & 44.4 & 38.5 & 47.4 \\
L-Seg~\cite{anonymous2022languagedriven} & Zero & ViT-L/16 & \textbf{61.3} & 63.6 & 43.1 & 41.0 & 52.3 \\
Ours & Zero & Res50 & 60.3 & \textbf{68.5} & \textbf{55.5} & \textbf{55.5} & \textbf{59.9} \\ \midrule
Ours & One & Res50 & 62.6 & 69.1 & 56.1 & 56.7 & 61.1 \\
Ours & Five & Res50 & \textbf{63.6} & \textbf{69.6} & \textbf{56.3} & \textbf{57.4} & \textbf{61.8} \\ \bottomrule
\end{tabular}}
\vspace{-.5em}
\caption{Comparisons with ZSS methods on $\text{Pascal}\textit{-}5^i$.}
\label{zeroshot}
\vspace{-0.5em}
\end{table}

\begin{table}[t]
\resizebox{\columnwidth}{!}{
\begin{tabular}{@{}l|ccccccc@{}}
\toprule
Recurrent Cell & $5^{0}$ & $5^{1}$ & $5^{2}$ & $5^{3}$ & Mean & Params & MACs \\ \midrule
None & 59.5 & 67.5 & 54.1 & 53.8 & 58.7 & \textbf{26.1M} & \textbf{7.1G} \\
Conv & 59.6 & 59.4 & \textbf{56.3} & 56.4 & 60.4 & 29.9M & 22.4G \\
Group Conv & \textbf{62.6} & \textbf{69.1} & 56.1 & \textbf{56.7} & \textbf{61.1} & 28.5M & 19.0G \\ \bottomrule
\end{tabular}}
\caption{Comparisons of different recurrent cell in IMR. We report the parameters without those of the shared CLIP-CAM, and the MACs of \textit{neck} and \textit{head} in one step.}
\vspace{-.5em}
\label{conv}
\end{table}

\begin{table}[t]
\centering
\scalebox{0.8}{
\begin{tabular}{@{}l|lcccc@{}}
\toprule
$\lambda$  & $5^{0}$ & $5^{1}$ & $5^{2}$ & $5^{3}$ & Mean \\ \midrule
None & 61.3 & 68.9 & 56.0 & 55.6 & 60.5 \\
Linear & 60.4 & 68.6 & 56.4 & 56.5 & 60.5 \\
Same & 62.6 & 69.1 & 56.1 & 56.7 & 61.1 \\ \bottomrule
\end{tabular}}
\vspace{-.5em}
\caption{Comparisons of different $\lambda_t$ in loss function}
\label{lambda}
\end{table}

\begin{table}[t]
\renewcommand\arraystretch{1.5}
\centering
\scalebox{0.8}{
\begin{tabular}{@{}l|llccc@{}}
\toprule
Prediction & $20^{0}$ & $20^{1}$ & $20^{2}$ & $20^{3}$ & Mean \\ \midrule
$\hat{M}_q^0$ & 17.5 & 18.8 & 21.0 & 19.5 & 19.2 \\
$\hat{M}_q^1$ & 36.3 & 41.7 & 41.4 & 41.3 & 40.2 \\
$\hat{M}_q^2$ & 38.6 & 43.2 & 42.1 & 42.9 & 41.7 \\
$\hat{M}_q^3$ & \textbf{39.5} & \textbf{43.8}& \textbf{42.4} & \textbf{44.1} & \textbf{42.4} \\ \bottomrule
\end{tabular}}
\vspace{-.5em}
\caption{Performance in the different steps of refinement on $\text{COCO}\textit{-}20^i$.}
\label{stage}
\end{table}

\begin{figure*}[t]
    \centering
    \vspace{-0.2cm}
    \includegraphics[width=\textwidth]{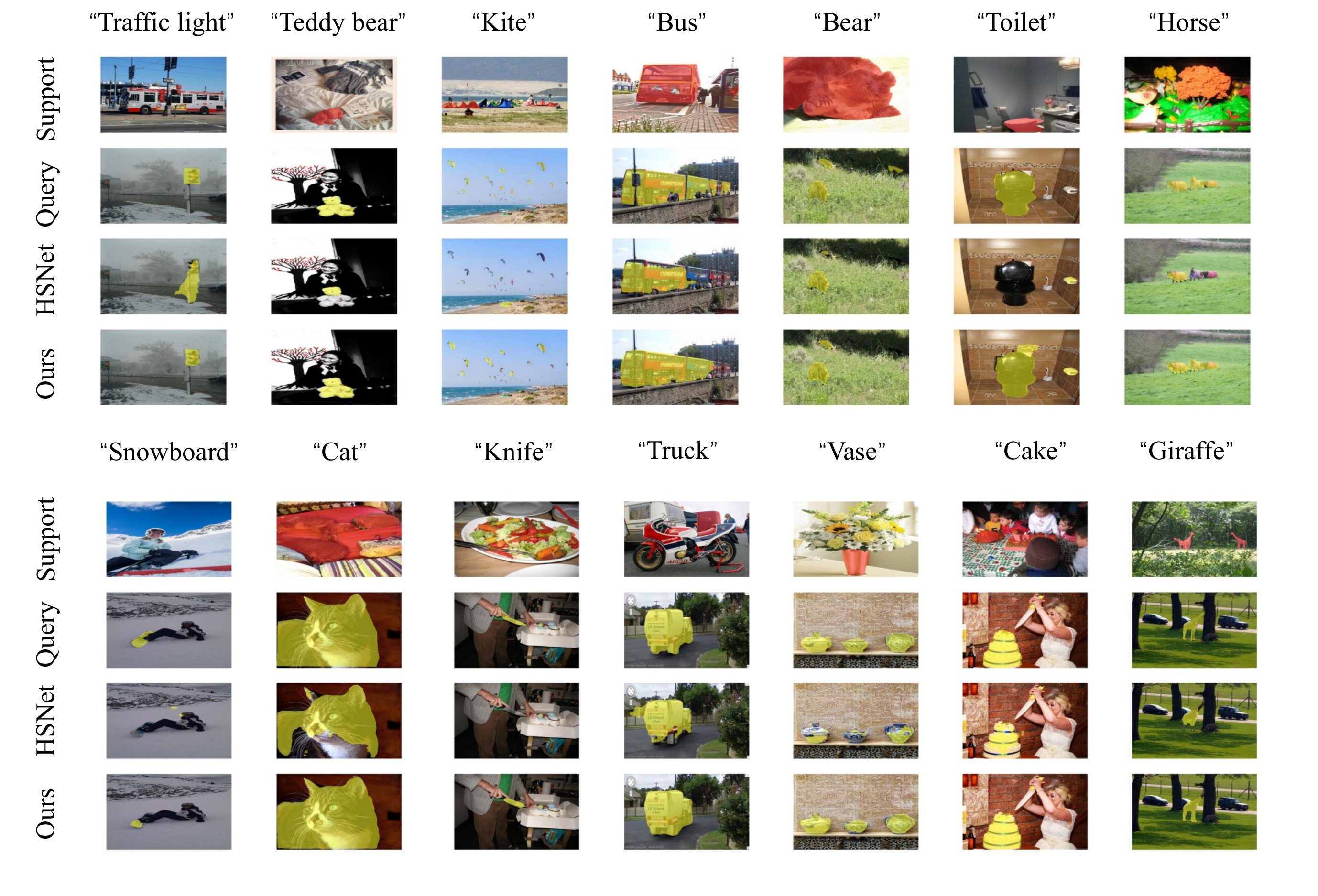}
    \vspace{-1.0cm}
    \caption{Qualitative comparison between our method and the state-of-the-art few-shot segmentation method HSNet. HSNet requires pixel-level support annotations to guide the segmentation, while we only need the image-level label text. However, we could obtain the performance on par with or even better than that of HSNet.}
    \label{coco1}
    \vspace{-0.2cm}
\end{figure*} 

\subsection{Architecture Design of Recurrent Cell in IMR}
In this subsection we explore the impact of different architecture design of recurrent cell in IMR. We set the method without recurrent cell as the baseline, where we simply utilize the pre-trained HSNet to first predict $\hat{M}^1_q$ with $\hat{M}^0_s$, then update $\hat{M}^2_s$ with $\hat{M}^1_q$, and finally segment $\hat{M}^3_q$ with the guidance of $\hat{M}^2_s$. Furthermore, we evaluate the regular convolutional GRU cell and the group convolutional GRU cell in IMR, where the $W_\text{conv1,1}, W_\text{conv1,2}, W_\text{conv2}$ of IMR is implemented with standard $3\times 3$ convolutions, and with $1\times 1$ convolutions followed by $3\times 3$ group convolutions with 16 groups, respectively.\par


\cref{conv} presents the comparison of performance as well as the whole parameters. Since IMR is incorporated into the \textit{neck} and \textit{head} of HSNet, we also compare the MACs of them in one step. It could be seen that both methods using the recurrent cell outperform the baseline with only a few extra parameters and acceptably more MACs. Furthermore, when utilizing group convolutions rather than regular ones as the recurrent cell in IMR, the higher performance is gained with the fewer parameters and MACs. The possible reason could be that the group convolutions encourage the feature regularization and thus improve the generalization ability. As a result, we adopt the group convolutions in the IMR.\par

\subsection{Ablation Study on the Hyperparameter}
In this subsection we study the sensitivity to hyperparameters of our method. As explained in our paper, we add the intermediate supervision after each iteration, and weight the corresponding loss by the hyper-parameter $\lambda_t$:
\begin{equation}
\mathcal{L}_\text{all} = \sum_{t=1}^N \lambda_t \cdot [\mathcal{L}_\text{seg}(\hat{M}_q^{t}, M_q) + w_s \cdot \mathcal{L}_\text{seg}(\hat{M}_s^{t}, M_s)].
\end{equation}
We further explore the impact of different $\lambda_t$. Supposing we optimize the estimation of support and query mask for $N$ iterations. Firstly we set $\lambda_t=0$ for all $t\neq N$ (denoted as \textit{None}), which means we do not utilize the intermediate supervision. We also set $\lambda_t=1$ to treat every iteration equally, denoted as \textit{Same}. Furthermore we try a simple linear strategy where $\lambda_t=\frac{t}{N}$, which is denoted as \textit{Linear}.\par

As reported in \cref{lambda}, training with different $\lambda_t$ leads to the similar results as the performance gap is less than 1\%. It turns out that our iterative strategy is robust to the setting of hyperparameters, which is friendly for practical applications.

\subsection{Effects of Iterative Refinement}
In this subsection we evaluate whether the quality of mask estimation is improving during iterative refinement. We conduct experiments on $\text{COCO}\textit{-}20^i$ as we iteratively optimize the mask thrice on that. \haohan{Specifically, we report the quality of the initial generated query mask $\hat{M_q^0}$ from Grad CAM, the intermediate predicted query mask $\hat{M_q^1},\hat{M_q^2}$ and the final $\hat{M_q^3}$, as shown in \cref{stage}.} It could be concluded that the predicted query mask is getting finer gradually with the iterative optimization, which further demonstrates the effectiveness of our iterative strategy.\par

\section{More Qualitative Results}
\cref{coco1} shows more qualitative results to compare our weakly supervised method with the state-of-the-art regular method HSNet. It could be seen that the performance of our method is on par with or even better than HSNet. When segmenting query images with multiple objects (such as kite, vase and horse), or tiny objects (such as traffic light, bear, snowboard and knife), our method shows strong superiority. Furthermore, when objects in query images look much different from in support images, our method is able to generate more complete mask (such as teddy bear, cat and cake).\par

\bibliographystyle{named}
\bibliography{ijcai22}